\documentclass{article}

    \PassOptionsToPackage{numbers, compress}{natbib}
 \usepackage[dblblindworkshop, final]{neurips_2026}
\workshoptitle{Representations for the Physical Sciences}
\usepackage[utf8]{inputenc} % allow utf-8 input
\usepackage[T1]{fontenc}    % use 8-bit T1 fonts
\usepackage{hyperref}       % hyperlinks
\usepackage{url}            % simple URL typesetting
\usepackage{booktabs}       % professional-quality tables
\usepackage{amsfonts}       % blackboard math symbols
\usepackage{nicefrac}       % compact symbols for 1/2, etc.
\usepackage{microtype}      % microtypography
\usepackage{xcolor}         % colors
\usepackage{mathtools}
\usepackage{graphicx}
\usepackage{subcaption}

\title{\(C_4\)-Equivariant Flow Matching on Anisotropic Power-Diagram Graphs for Microstructure Generation} % align more wiht workghsopt - ML with Imperfect Scientific Models

\author{
  Dawid Lipinski, Jixiang Qing, Henry Moss\\
  School of Mathematical Sciences\\
  Lancaster University\\
  Lancaster, LA1 4YF, United Kingdom\\
  \texttt{\{d.a.lipinski, j.qing, henry.moss\}@lancaster.ac.uk}
}

\begin{document}

\maketitle

\begin{abstract}
  Acquiring realistic microstructure data through Electron Backscatter Diffraction (EBSD) is costly and time consuming, often relying on specialised equipment. As microstructures strongly influence material properties, generating realistic samples is essential for modelling the behaviour of polycrystalline materials. We introduce a generative model for synthesising realistic polycrystalline microstructures using flow matching and graph neural networks. By representing microstructures as anisotropic power diagrams, our model learns a compact geometric parametrisation and can render generated samples at arbitrary pixel resolution. A \(C_4\)-equivariant architecture incorporates rotational symmetry directly into the model, ensuring that rotations of the input noise produce corresponding rotations of the generated microstructure. We also demonstrate how training-free guidance can be used to generate complex microstructures, based on user defined objective function. In particular, we generate microstructures resembling a copper weld, cast metal slab, 3D-printed stainless steel and heterogeneous lamella titanium.
\end{abstract}

% Things I will try to add:
% - fine tuning on reald ata
% - training only on real (but limited) data
% - more correlation/experimental metrics?
% - 

\section{Introduction}
\label{Intro}
The properties of polycrystalline materials depend strongly on the geometry and spatial arrangement of their grains. Representative microstructures are therefore essential for modelling material behaviour and studying structure–property relationships. Although Electron Backscatter Diffraction (EBSD) \cite{schwartz_electron_2009} provides detailed experimental measurements, acquiring large and diverse datasets is costly and time-consuming. Synthetic microstructures are consequently widely used in computational materials science, with approaches ranging from tessellation-based methods to learned generative models \cite{hitti_direct_2011, buze_anisotropic_2024, labady_infinityebsd_2025, sekwao_end--end_2025}. Learning directly from EBSD-derived images remains challenging because the available data are often limited, noisy, and too large to process as complete samples. Moreover, many existing approaches offer limited control over global characteristics of the synthesised samples, such as collective grain alignment or spatial patterns extending across the entire microstructure \cite{buze_anisotropic_2024, labady_infinityebsd_2025}.

Voronoi and power diagrams provide compact geometric representations of polycrystalline microstructures, but their ability to represent anisotropic grains and curved boundaries is limited \cite{hitti_direct_2011}. Anisotropic power diagrams (APDs) augment ordinary power diagrams with grain-specific anisotropy, allowing directionally dependent grain geometries while retaining a compact parametric representation, while recent GPU-accelerated methods enable them to be fitted and generated efficiently \cite{buze_anisotropic_2024}. Existing APD generation methods prescribe selected grain statistics but do not learn a generative distribution over complete grain configurations or directly model spatial dependencies \cite{buze_anisotropic_2024}. Furthermore, control over global characteristics is typically limited to statistics specified a priori by the sampling procedure.

\begin{figure}
    \centering

    \begin{subfigure}[b]{0.49\textwidth}
        \centering
        \includegraphics[width=\textwidth]{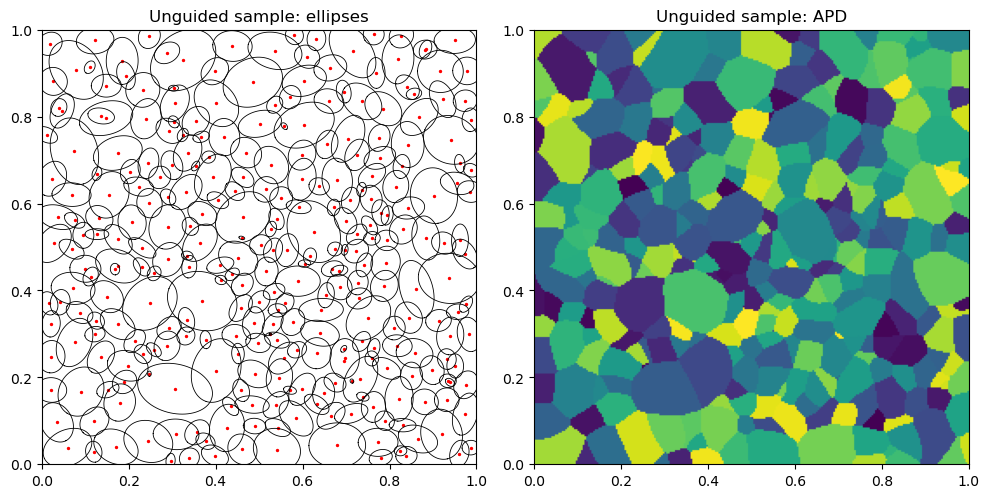}
        % \caption{Unguided generation.}
        % \label{fig:APD_unguided}
    \end{subfigure}
    \hfill
    \begin{subfigure}[b]{0.49\textwidth}
        \centering
        \includegraphics[width=\textwidth]{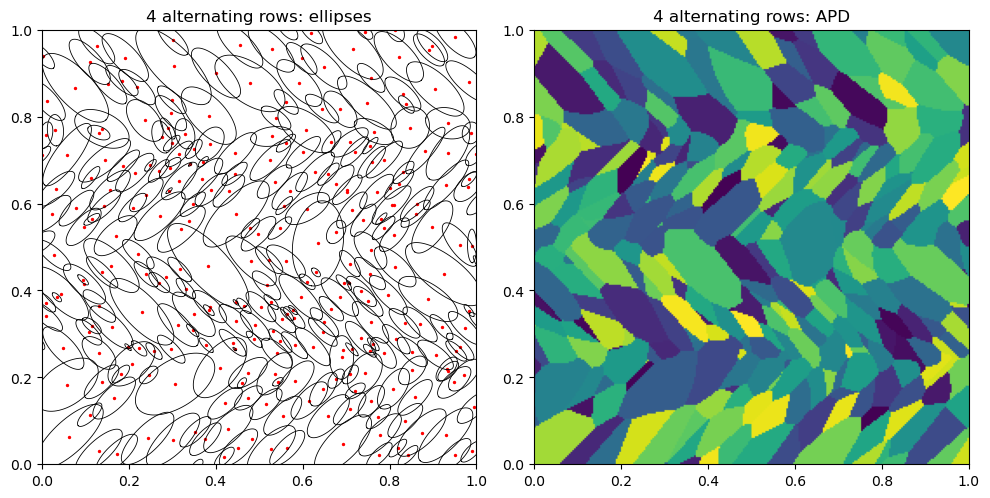}
        % \caption{Guided generation.}
        % \label{fig:APD_guided}
    \end{subfigure}

    \caption{
        Unguided (left) and guided (right) samples generated by our model from the same initial noise. Each panel shows the ellipse representation on the left and the
        corresponding APD tessellation on the right. Guidance encourages four
        horizontal bands with alternating grain orientations of
        \(+45^\circ\) and \(-45^\circ\), resembling a microstructure pattern in 3D-printed stainless steel.
    }
    \label{fig:APD_example}
\end{figure}

Our main contributions are:
\begin{enumerate}
    \item We formulate polycrystalline microstructure generation as flow matching over the compact geometric parameters of an anisotropic power diagram. Each grain is represented by its position in \([0,1]^2\) and an ellipse-based parametrisation of its anisotropy; see Figure~\ref{fig:APD_example} for illustration. This resolution-independent representation supports variable grain counts and allows generated microstructures to be rendered at arbitrary pixel resolution.
    \item We introduce a graph-based, \(C_4\)-equivariant architecture that models dependencies between grains and ensures consistent generative behaviour under rotations by multiples of \(90^\circ\).
    \item We introduce a training-free guidance framework for controlling generated microstructures through user-defined differentiable objectives. Without retraining, the framework can impose complex, spatially varying global structure, enabling the generation of diverse morphologies, including examples qualitatively inspired by copper welds, cast metal slabs, 3D-printed stainless steel and heterogeneous lamella titanium; see Figure~\ref{fig:guidance_examples}. The code can be found at: \url{https://github.com/DawidLipin/Flow_APDs}
\end{enumerate}

\section{Background}
\label{sec:Background}

\textbf{Anisotropic power diagrams.}
We define an \emph{anisotropy matrix} \(A\in\mathbb{R}^{2\times 2}\) as a symmetric positive-definite matrix, inducing the anisotropic norm \(\lVert x\rVert_A \coloneqq \sqrt{x^\top A x}.\)
Let \(D=\{(x_i,A_i,w_i)\}_{i=1}^N,\) where \(x_i\in[0,1]^2\) is a seed point, \(A_i\in\mathbb{R}^{2\times2}\) is an anisotropy matrix, \(w_i\in\mathbb{R}\) is a scalar weight, and \(N\) is the number of regions. For each \(i\in\{1,\ldots,N\}\), we define
\[
    L_i \coloneqq
    \left\{
        x\in[0,1]^2 :
        \lVert x-x_i\rVert_{A_i}^2 - w_i
        \leq
        \lVert x-x_j\rVert_{A_j}^2 - w_j
        \quad
        \forall j\in\{1,\ldots,N\}
    \right\}.
\]
The regions \(\{L_i\}_{i=1}^N\) form a tessellation of \([0,1]^2\), which we refer to as an \emph{anisotropic power diagram} (APD). See Figure~\ref{fig:APD_example} for an illustration.
Intuitively, each triple \((x_i,A_i,w_i)\) determines one region of the tessellation. The seed \(x_i\) specifies its location, while \(A_i\) determines the associated anisotropic distance. In particular, the unit ball of \(\lVert\cdot\rVert_{A_i}\) is an ellipse centred at \(x_i\). The weight \(w_i\) controls the relative influence of the seed and therefore affects the size of the resulting region.\\
We associate each region with a target area \(v_i>0\), where \(\sum_{i=1}^N v_i = 1.\) An APD is \emph{optimal} with respect to the target areas
\(V=\{v_i\}_{i=1}^N\) if \(|L_i| = v_i\) for all \(i\), where \(|L_i|\) denotes the area of region \(L_i\). For fixed seed positions \(x_i\), anisotropy matrices \(A_i\), and target areas \(v_i\), the weights \(w_i\) can be obtained by solving an optimal transport problem \cite{buze_anisotropic_2024} so that the regions satisfy the prescribed areas. This construction allows arbitrary target tessellations to be approximated by optimising the APD parameters \(\{(x_i,A_i,w_i)\}_{i=1}^N\) such that the induced regions \(L_i\) match the target regions.

\textbf{Flow matching}. Flow matching trains a continuous normalising flow to transport samples from a simple source distribution \(p_0\), typically Gaussian noise, to a target data distribution \(p_1\) \cite{lipman_flow_2023}. The transformation is defined by the ordinary differential equation
\[
    \frac{\mathrm{d}\mathbf{z}_t}{\mathrm{d}t}
    = v_\theta(\mathbf{z}_t,t),
    \qquad \mathbf{z}_0\sim p_0,
\]
where \(v_\theta\) is a vector field constructed using a neural network. For the linear conditional path \(\mathbf{z}_t=(1-t)\mathbf{z}_0+t\mathbf{z}_1\) with \(t\sim\mathcal{U}(0,1) \) and the corresponding target velocity given by \(\mathbf{z}_1-\mathbf{z}_0\), this gives the simulation-free training objective
\[
    \mathcal{L}_{\mathrm{FM}}(\theta)
    =
    \mathbb{E}\!\left[
    \left\|
    v_\theta(\mathbf{z}_t,t)
    -(\mathbf{z}_1-\mathbf{z}_0)
    \right\|_2^2
    \right].
\]
Samples are generated by numerically integrating the learned vector field from \(t=0\) to \(t=1\). In our setting, \(\mathbf{z}_t\) contains the APD parameters of all grains and \(v_\theta\) is parametrised by a graph neural network.

\textbf{Training-free guidance.} Following gradient-based guidance methods for diffusion and flow models \cite{chung_diffusion_2022, ye_tfg_2024}, we bias generation toward a differentiable objective \(\mathcal{J}\) by instead sampling
\[
    p_1^{\mathcal{J}}(\mathbf{z}_1)
    \propto
    p_1(\mathbf{z}_1)
    \exp\!\left[-\beta\mathcal{J}(\mathbf{z}_1)\right],
\]
where \(\beta\) controls the guidance strength. At time \(t\), we estimate the terminal state by \(\widehat{\mathbf{z}}_1 = \mathbf{z}_t+(1-t)v_\theta(\mathbf{z}_t,t)\) and modify the velocity as
\[
    \widetilde{v}_\theta(\mathbf{z}_t,t)
    =
    v_\theta(\mathbf{z}_t,t)
    -
    s(t) \beta
    \nabla_{\mathbf{z}_t}
    \mathcal{J}(\widehat{\mathbf{z}}_1),
\]
where \(s(t)\) is a time-dependent schedule. The model parameters remain fixed, enabling training-free control through arbitrary differentiable APD objectives.

\section{Equivariant flow matching on APDs}
\label{sec:model}

Directly modelling EBSD images is computationally expensive, as individual samples may contain millions of pixels, while generating realistic microstructures requires preserving their underlying geometric structure. We address these challenges by representing microstructures compactly as APDs and using an equivariant graph architecture that enforces consistent behaviour under rotations.

\textbf{Microstructure representation.} We represent a microstructure with \(N\) grains by \(\mathbf{z} = \{(x_i,a_i,w_i)\}_{i=1}^{N},\) where \(x_i\in[0,1]^2\) is the seed position, \(w_i\in\mathbb{R}\) its power weight, and \(a_i\in\mathbb{R}^2\) parametrises its anisotropy. This compact APD representation is independent of raster resolution and regularises small pixel-level segmentation artefacts. Details of the reparametrisation \(A_i\to a_i\) are given in Appendix~\ref{app:anisotropy}.

\textbf{Equivariant graph architecture.}
We represent each APD as a graph whose nodes are \(\{(x_i,a_i,w_i)\}_{i=1}^{N},\) and whose directed edges \(e_{i,j}\in\mathbb{R}^{15}\)  connect each source \(x_i\) node to its \(k\) nearest neighbours \(x_j\). The invariant edge features supplied to the message-passing network are
\[
\phi_{ij}=
\big[
 \|d_{ij}\|^2,\,
 w_i,w_j,\,
 \|a_i\|^2,\|a_j\|^2,\,
 a_i^\top a_j,\,
 a_i\times a_j,\,
 a_i^\top\rho(d_{ij}),\,
 a_j^\top\rho(d_{ij}),\,
 b(x_i),b(x_j)
\big]\in\mathbb R^{15},
\]
where \(d_{ij} = x_i - x_j\) and, for \(u=(u_1,u_2)\) and \(v=(v_1,v_2)\), we define
\(
u\times v=u_1v_2-u_2v_1\), \(
\rho(u)=(u_1^2-u_2^2,\,2u_1u_2),
\)
and
\(
b(u)=(u_1^2u_2^2,\,u_1^2+u_2^2,\,u_1^4+u_2^4)
\)
encodes position relative to the square boundary while remaining
\(C_4\)-invariant. These features are processed by message-passing layers conditioned on the flow time \(t\), with aggregation ensuring permutation equivariance. The output head predicts \(C_4\)-invariant scalar coefficients multiplying geometric bases that transform equivariantly under \(90^\circ\) rotations; see Appendix~\ref{app:equivariant-updates} for details. Hence, for any \(90^\circ\) rotation \(R\), we have \(v_\theta(R\mathbf{z},t)=R\,v_\theta(\mathbf{z},t),\) so rotating the initial state produces the correspondingly rotated generated APD.

\section{Experiments}
\label{sec:experiments}

\textbf{Experimental setup.} We trained a model with three message-passing layers, a latent dimension of \(256\), and \(k=16\) nearest neighbours. The training set contained \(12{,}000\) synthetic APDs with between \(100\) and \(500\) grains, generated using PyAPD \cite{buze_anisotropic_2024}. The model was trained for 100 epochs, using the AdamW optimiser \cite{loshchilov_decoupled_2018}. The code is available at: \url{https://github.com/DawidLipin/Flow_APDs}

\textbf{Training-free guidance.} Following the convention in Section~\ref{sec:Background}, for each APD channel
\(c\in\{x,a,w\}\), we use
\[
 \widetilde v_{\theta,c}
 =v_{\theta,c}
 -s(t)\sum_k \lambda_{k,c}
 \frac{\nabla_{z_{t,c}}\mathcal J_k(\widehat{\mathbf z}_1)}
 {\max\!\left(\operatorname{RMS}
 (\nabla_{z_{t,c}}\mathcal J_k),\epsilon\right)},
\]
where RMS stands for root mean square, \(\epsilon=10^{-8}\) ensures numerical stability, and \(\lambda_{k,c}\) independently controls the effect of objective \(k\) on generator positions, anisotropies, and power weights. The gradients are obtained by backpropagating both through the objective and \(v_\theta\). Normalising each gradient separately prevents its numerical scale from determining its influence. We use a cosine ramp \(s(t)\), with no guidance over the first \(10\%\) of the trajectory and recompute the neighbourhood graph as the generator positions evolve.

\begin{figure}
    \centering
    \includegraphics[width=0.99\textwidth]{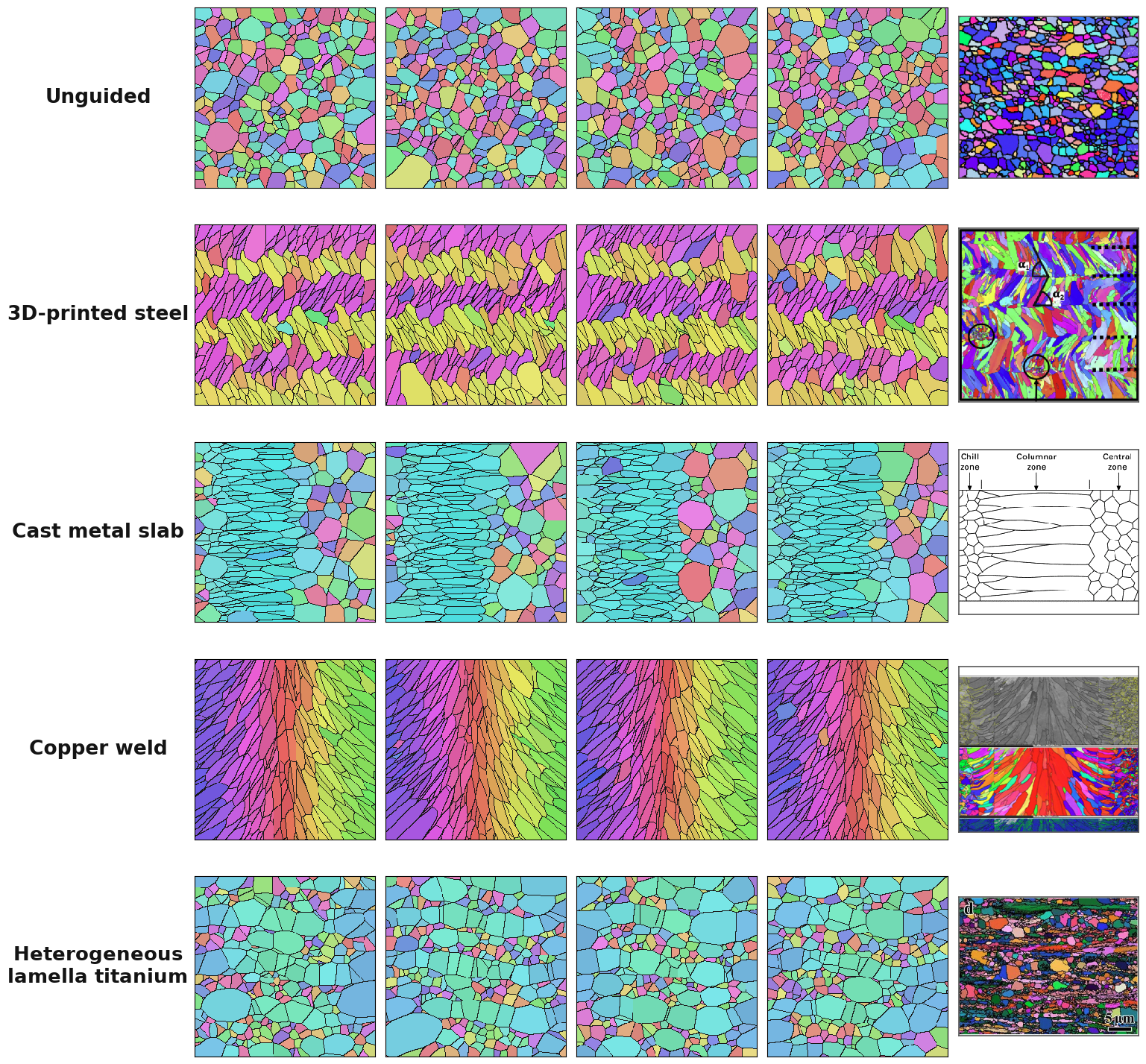}
    \caption{Each row shows training-free guidance of synthetic polycrystalline microstructures. The first four columns show independent samples generated by the same pretrained model, while the rightmost column presents representative experimental microstructures for qualitative comparison. The real examples are taken from (top to bottom): \citet{buze_anisotropic_2024, balit_digital_2020, reed-hill_physical_1973, savolainen_localization_2011, wu_heterogeneous_2015}}
    \label{fig:guidance_examples}
\end{figure}

The four columns in Figure~\ref{fig:guidance_examples} are independent
samples from the same pretrained model, using various guidance functions. Other objectives can also be used depending on the application. The
3D-printed example imposes elongated grains with alternating layer-wise
directions; the cast slab combines spatially varying shape, relative grain
weight, and generator-density targets for chill, columnar, and central zones; the weld objective prescribes a smoothly varying inward growth field; and the heterogeneous-lamella objective places a high-weight coarse population in wavy bands while the surrounding smaller grains remain approximately equiaxed. These objectives constrain only selected statistics, leaving the pretrained flow to determine the remaining local geometry. The guidance objectives from top to bottom are

\begin{align*}
\mathcal J_{\mathrm{layer}}
&=
\frac{\sum_i\text{sg}(|\ell_i|)
\|a_i-a^\star(\theta_i^{\mathrm{lay}},5.5)\|_2^2}
{\sum_i\text{sg}(|\ell_i|)},
\quad
\ell_i=\tanh(12\sin(6\pi y_i)),
\quad
\theta_i^{\mathrm{lay}}=90^\circ+35^\circ\ell_i,\\
\boldsymbol{\mathcal J}_{\mathrm{cast}}
&=
\left(
\left\langle\|a_i-a_i^{\mathrm{cast}}\|_2^2\right\rangle,\;
\operatorname{CMSE}(w,q^{\mathrm{cast}}),\;
\left\langle[x_{(i)}-\tau_i]^2\right\rangle
\right),\\
\mathcal J_{\mathrm{weld}}
&=
\left\langle
\|a_i-a^\star(\theta_i^{\mathrm{weld}},5.5)\|_2^2
\right\rangle,
\quad
\theta_i^{\mathrm{weld}}
=
\operatorname{atan2}(0.62,1.9(0.5-x_i)),\\
\boldsymbol{\mathcal J}_{\mathrm{hl}}
&=
\left(
-\langle\log p_i\rangle_{\mathcal C}
-\langle\log(1-p_i)\rangle_{\mathcal F},\;
\operatorname{SMSE}(w,q^{\mathrm{hl}}),\;
\left\langle\|a_i-a_i^{\mathrm{band}}\|_2^2\right\rangle_{\mathcal C}
+
\left\langle\|\sigma_a a_i\|_2^2\right\rangle_{\mathcal F}
\right).
\end{align*}
where for compactness, we write
\(
a^\star(\theta,r)
=
-\frac{\log r}{\sigma_a}
\bigl(\cos(2\theta),\sin(2\theta)\bigr),
\)
\(
\langle f_i\rangle_{\mathcal S}
=
\frac1{|\mathcal S|}\sum_{i\in\mathcal S}f_i,
\)
sg for the stop-gradient,
and CMSE and SMSE for the centred and sorted MSE respectively.
For the cast objective, \(a_i^{\mathrm{cast}}\), \(q_i^{\mathrm{cast}}\), and \(\tau_i\)
prescribe the cast-slab shape, relative-weight, and generator-density
profiles. For the heterogeneous-lamella objective,
\(\mathcal C\) and \(\mathcal F\) are the coarse and fine populations,
\(p_i\) is the band-membership target, \(q_i^{\mathrm{hl}}\) is a two-level relative-weight target, and \(a_i^{\mathrm{band}}\) aligns coarse grains with the local band tangent.
Components of vector-valued objectives are differentiated and RMS-normalised separately.
Their precise definitions and all guidance strengths are given in
Appendix~\ref{app:guidance-objectives}. 
We selected the target parameters and guidance strengths manually to obtain qualitative agreement with the representative experimental microstructures. These values need not be optimal and, for a given application, may instead be specified using domain knowledge, tuned against an application-specific validation criterion, or estimated from real data.

\section{Discussion}
\label{Discussion}
The model is currently trained only on synthetic APDs, so fidelity to
higher-order experimental statistics remains unvalidated; strong or conflicting
guidance may also leave the learned distribution. Future work will aim to evaluate and
fine-tune the model on EBSD data, incorporate crystallographic orientation, and extend the
representation to three-dimensional microstructures.
Exploring alternative guidance objectives, particularly those based on physics-informed loss functions, is also a promising direction for future research.

\bibliographystyle{plainnat}
\bibliography{APD_Diffusion_NeurIPS}

\appendix

\section{Guidance objectives and sampling details}
\label{app:guidance-objectives}

This section gives the complete objectives and sampling settings used to
produce Figure~\ref{fig:guidance_examples}. Experimental reference images are
used only for qualitative comparison and do not enter any objective.

\paragraph{Coordinates and common notation.}
The flow operates in normalised coordinates. To distinguish model and physical
coordinates, let $\widetilde x_i\in\mathbb R^2$ denote a normalised generator
position and let
\[
 (x_i,y_i)=\frac{\widetilde x_i}{2}+\left(\frac12,\frac12\right)
\]
be the corresponding position in the physical coordinate system. Data
endpoints lie in $[0,1]^2$, but this affine map is applied without clipping to
intermediate terminal estimates, which can temporarily leave the domain. In
the equations below, all spatial target fields use
$(x_i,y_i)$, whereas $a_i\in\mathbb R^2$ and $w_i\in\mathbb R$ denote the
normalised anisotropy and weight channels. The physical log-anisotropy is
$\sigma_a a_i$, where $\sigma_a$ is the scalar anisotropy scale measured on
the training set. The normalised spin--2 encoding of a grain whose visible
long axis has angle $\theta$ and axis ratio $r\geq1$ is
\[
 a^\star(\theta,r)
 =-\frac{\log r}{\sigma_a}
 \bigl(\cos(2\theta),\sin(2\theta)\bigr).
\]
Thus $a^\star(\theta+\pi,r)=a^\star(\theta,r)$ and
$a^\star(\theta,1)=0$, as required for unoriented grain axes and equiaxed
grains, respectively. Angles are in radians unless a degree symbol is shown.

We use
\[
 \langle f_i\rangle_{\mathcal S}
 =\frac{1}{|\mathcal S|}\sum_{i\in\mathcal S}f_i,\qquad
 \langle f_i\rangle=\frac1N\sum_{i=1}^N f_i.
\]
For a non-negative pairwise kernel $K_{ij}$, we also write
\[
 \langle H_{ij}\rangle_K
 =\frac{\sum_{i\ne j}K_{ij}H_{ij}}
 {\sum_{i\ne j}K_{ij}}.
\]
For scalar vectors $u,v\in\mathbb R^N$, define
\begin{align*}
 \operatorname{CMSE}(u,v)
 &=\frac1N\sum_{i=1}^N
 \left[(u_i-\bar u)-(v_i-\bar v)\right]^2,\\
 \operatorname{SMSE}(u,v)
 &=\frac1N\sum_{i=1}^N
 \left[(u-\bar u)_{(i)}-(v-\bar v)_{(i)}\right]^2,
\end{align*}
where $\bar u=N^{-1}\sum_i u_i$, and $(\cdot)_{(i)}$ denotes the
$i$th order statistic. CMSE constrains relative APD weights without fixing
their irrelevant additive gauge; SMSE also removes dependence on grain labels.
For the two anisotropy components, the implementation uses the componentwise
mean-squared error
\[
 \operatorname{MSE}_a(u,v;\mathcal S)
 =\frac{1}{2|\mathcal S|}\sum_{i\in\mathcal S}\|u_i-v_i\|_2^2.
\]
The factor $1/2$ simply averages over the two anisotropy components. Global
positive rescalings of an individual objective have no effect after the
gradient normalisation described below.

\paragraph{Endpoint guidance and numerical integration.}
At time $t$, the model gives the first-order terminal estimate
\[
 \widehat{\mathbf z}_1(\mathbf z_t)
 =\mathbf z_t+(1-t)v_\theta(\mathbf z_t,t).
\]
For objective $\mathcal J_k$ and state channel
$c\in\{x,a,w\}$, let
\[
 g_{k,c}
 =\nabla_{z_{t,c}}
 \mathcal J_k\!\left(\widehat{\mathbf z}_1(\mathbf z_t)\right),
 \qquad
 \operatorname{RMS}(g_{k,c})
 =\left(\frac{1}{D_c}\sum_{j=1}^{D_c}g_{k,c,j}^2\right)^{1/2},
\]
where $D_c$ is the number of scalar entries in channel $c$. Gradients pass
through both the objective and $v_\theta$ in the terminal estimate. The guided
velocity is
\[
 \widetilde v_{\theta,c}
 =v_{\theta,c}
 -s(t)\sum_k\lambda_{k,c}
 \odot
 \frac{g_{k,c}}
 {\max\{\operatorname{RMS}(g_{k,c}),10^{-8}\}},
\]
where $\odot$ denotes componentwise multiplication. A scalar
$\lambda_{k,c}$ is broadcast over a channel, while
$\lambda_{k,x}=(\lambda_x^{(1)},\lambda_x^{(2)})$ permits different
horizontal and vertical position strengths. Each objective--channel pair is
RMS-normalised separately before the corrections are summed. The scalar
strength $\beta$ in Section~\ref{sec:Background} is therefore absorbed into
the experimentally used $\lambda_{k,c}$.
Table~\ref{tab:guidance-strengths} provides the specific guidance strengths used.

The schedule has a $10\%$ unguided warm-up followed by a cosine ramp,
\[
 s(t)=
 \begin{cases}
 0, & 0\leq t\leq t_0,\\[2pt]
 \dfrac12\left[1-\cos\!\left(\pi\dfrac{t-t_0}{1-t_0}\right)\right],
 & t_0<t\leq1,
 \end{cases}
 \qquad t_0=0.10.
\]
All samples in the figure use $N=300$ generators, $k=16$ nearest neighbours,
and $K=60$ Euler steps with $t_n=n/K$ and
\[
 \mathbf z_{n+1}
 =\mathbf z_n+\frac1K\widetilde v_\theta(\mathbf z_n,t_n).
\]
Each normalised channel of the base state is sampled independently from a
standard Gaussian. The four generated columns use random seeds $117$, $237$,
$377$, and $527$, respectively; within a column, every row starts from the
same base state. The $k$-nearest-neighbour graph is recomputed after
every step as generator
positions evolve. The updated state is detached before the next step, so the
method differentiates through the current terminal estimate but not through
the complete integration history. Neighbour selection is discrete: gradients
pass through the network evaluated on the current edge set, but not through the
choice of the $k$ nearest neighbours. The unguided row uses an empty objective
set, for which $\widetilde v_\theta=v_\theta$.

\paragraph{Layered 3D-printed structure.}
The smooth layer signal, target direction, and confidence are
\[
 \ell_i=\tanh\!\left(12\sin(6\pi y_i)\right),\qquad
 \theta_i^{\mathrm{lay}}=90^\circ+35^\circ\ell_i,
 \qquad c_i=\operatorname{sg}(|\ell_i|),
\]
where $\operatorname{sg}$ denotes stop-gradient. The signal alternates sign
across six horizontal layers. The confidence removes the ambiguous orientation
at layer interfaces, while detaching it prevents grains from reducing the loss
merely by moving towards an interface. The objective is
\[
 \mathcal J_{\mathrm{layer}}
 =\frac{\sum_{i=1}^N c_i\,
 \tfrac12\|a_i-a^\star(\theta_i^{\mathrm{lay}},5.5)\|_2^2}
 {\sum_{i=1}^N c_i}.
\]

\paragraph{Cast slab.}
Let $b_{\mathrm{ch}}=0.11$ and $b_{\mathrm{co}}=0.58$ denote the ends of
the chill and columnar regions. Smooth transitions between the chill,
columnar, and core zones are constructed from
\[
\begin{aligned}
 u_i&=\operatorname{sigmoid}\!\left(45(x_i-b_{\mathrm{ch}})\right),
 &v_i&=\operatorname{sigmoid}\!\left(45(x_i-b_{\mathrm{co}})\right),\\
 \rho_i&=\operatorname{clip}\!\left(
 \frac{x_i-b_{\mathrm{ch}}}{b_{\mathrm{co}}-b_{\mathrm{ch}}},0,1
 \right).&&
\end{aligned}
\]
Writing
\[
 D_i=(1-u_i)+u_i(1-v_i)+v_i,
\]
the normalised zone memberships are
\[
 (m_i^{\mathrm{ch}},m_i^{\mathrm{col}},m_i^{\mathrm{core}})
 =\frac{(1-u_i,\;u_i(1-v_i),\;v_i)}{D_i}.
\]
The chill and core targets are equiaxed. In the columnar region the target is
horizontal and its axis ratio increases from $2.2$ to $9.2$:
\[
 a_i^{\mathrm{cast}}
 =m_i^{\mathrm{col}}a^\star(0^\circ,2.2+7\rho_i).
\]
The relative-weight target is
\[
 q_i^{\mathrm{cast}}
 =-1.2m_i^{\mathrm{ch}}
 +(-0.2+\rho_i)m_i^{\mathrm{col}}
 +1.4m_i^{\mathrm{core}}.
\]
Finally, the generator-density target assigns fractions
$f_{\mathrm{ch}}=0.34$, $f_{\mathrm{col}}=0.48$, and
$f_{\mathrm{core}}=0.18$ to the three zones. Let
\[
 s_i=\frac{i-\tfrac12}{N},\qquad
 f_{\mathrm{cc}}=f_{\mathrm{ch}}+f_{\mathrm{col}}=0.82,
\]
and define the piecewise-uniform quantile function
\[
 \tau(s)=
 \begin{cases}
 b_{\mathrm{ch}}\,s/f_{\mathrm{ch}},
 &0\leq s<f_{\mathrm{ch}},\\[2pt]
 b_{\mathrm{ch}}+
 (b_{\mathrm{co}}-b_{\mathrm{ch}})
 (s-f_{\mathrm{ch}})/f_{\mathrm{col}},
 &f_{\mathrm{ch}}\leq s<f_{\mathrm{cc}},\\[2pt]
 b_{\mathrm{co}}+(1-b_{\mathrm{co}})
 (s-f_{\mathrm{cc}})/f_{\mathrm{core}},
 &f_{\mathrm{cc}}\leq s\leq1.
 \end{cases}
\]
With $x_{(i)}$ the sorted physical horizontal coordinates, the three
independently guided cast objectives are
\[
 \boldsymbol{\mathcal J}_{\mathrm{cast}}
 =
 \left(
 \frac1{2N}\sum_{i=1}^N
 \|a_i-a_i^{\mathrm{cast}}\|_2^2,\;
 \operatorname{CMSE}(w,q^{\mathrm{cast}}),\;
 \frac1N\sum_{i=1}^N[x_{(i)}-\tau(s_i)]^2
 \right).
\]
Because $a_i^{\mathrm{cast}}$ and $q_i^{\mathrm{cast}}$ depend on position,
their objectives may guide both position and their corresponding APD channel.

\paragraph{Weld structure.}
The target vector field and its unoriented long-axis angle are
\[
 d_i=\bigl(1.9(0.5-x_i),0.62\bigr),\qquad
 \theta_i^{\mathrm{weld}}
 =\operatorname{atan2}\!\left(0.62,1.9(0.5-x_i)\right).
\]
The field points obliquely inward from either side and becomes vertical at the
centreline. Only its angle is needed by the objective:
\[
 \mathcal J_{\mathrm{weld}}
 =\frac1{2N}\sum_{i=1}^N
 \left\|a_i-a^\star(\theta_i^{\mathrm{weld}},5.5)\right\|_2^2.
\]
Although $d_i$ makes the construction intuitive, it is used mathematically
through $\theta_i^{\mathrm{weld}}=\arg(d_i)$.

\paragraph{Heterogeneous-lamella structure.}
Let
\[
 N_{\mathcal C}=\operatorname{round}(0.22N),\qquad
 N_{\mathcal F}=N-N_{\mathcal C}.
\]
At each endpoint estimate, $\mathcal C$ contains the
$N_{\mathcal C}$ largest centred weights and
$\mathcal F=\{1,\ldots,N\}\setminus\mathcal C$. The partition is discrete;
gradients are computed conditional on the roles selected at that step.

Four wavy bands are represented by
\begin{align*}
 h(x)&=0.035\sin(2\pi x),\\
 p_i&=\operatorname{sigmoid}\!\left[
 18\left(
 \cos\!\left(8\pi[y_i-h(x_i)]\right)-\cos(0.22\pi)
 \right)\right],\\
 \theta_i^{\mathrm{band}}
 &=\operatorname{atan2}\!\left(h'(x_i),1\right)
 =\operatorname{atan2}\!\left(0.07\pi\cos(2\pi x_i),1\right).
\end{align*}
The threshold $\cos(0.22\pi)$ makes the high-probability portion of each
period approximately match the $22\%$ coarse fraction. For numerical stability
the logarithms use
$\widehat p_i=\operatorname{clip}(p_i,10^{-6},1-10^{-6})$.

The zero-mean two-level relative-weight target is
\[
 q_i^{\mathrm{hl}}=
 \begin{cases}
 1.55, & i\in\mathcal C,\\[2pt]
 -1.55\,N_{\mathcal C}/N_{\mathcal F}, & i\in\mathcal F.
 \end{cases}
\]
The topology, relative-weight, and shape objectives are
\begin{align*}
 \mathcal J_{\mathrm{hl,top}}
 &=-\langle\log\widehat p_i\rangle_{\mathcal C}
   -\langle\log(1-\widehat p_i)\rangle_{\mathcal F},\\
 \mathcal J_{\mathrm{hl,weight}}
 &=\operatorname{SMSE}(w,q^{\mathrm{hl}}),\\
 \mathcal J_{\mathrm{hl,shape}}
 &=\frac1{2N_{\mathcal C}}\sum_{i\in\mathcal C}
 \left\|a_i-a^\star(\theta_i^{\mathrm{band}},1.6)\right\|_2^2
 +\frac1{N_{\mathcal F}}\sum_{i\in\mathcal F}
 \|\sigma_a a_i\|_2^2.
\end{align*}
The cross-entropy places the current coarse population in the bands, SMSE
maintains separated coarse and fine weight populations without fixing grain
labels, and the shape term weakly aligns coarse grains with the local band
tangent while driving fine grains towards equiaxed shapes.

\begin{table}[t]
\centering
\small
\begin{tabular}{lccc}
\toprule
Objective $\mathcal J_k$ & $\lambda_{k,x}$ & $\lambda_{k,a}$ & $\lambda_{k,w}$ \\
\midrule
Layer orientation & $(0,0.03)$ & $18$ & $0$ \\
Cast shape & $(0.02,0)$ & $16$ & $0$ \\
Cast relative weight & $(0.02,0)$ & $0$ & $0.30$ \\
Cast generator density & $(0.25,0)$ & $0$ & $0$ \\
Weld orientation & $0.02$ & $18$ & $0$ \\
Lamella topology & $(0.04,0.26)$ & $0$ & $0$ \\
Lamella shape & $0$ & $7$ & $0$ \\
Lamella relative weight & $0$ & $0$ & $0.36$ \\
\bottomrule
\end{tabular}
\caption{Guidance strengths used for Figure~\ref{fig:guidance_examples}.
An ordered pair gives horizontal and vertical position strengths; a scalar
position strength is applied to both components. Since each gradient is
RMS-normalised before scaling, the values specify correction magnitudes in
normalised model coordinates.}
\label{tab:guidance-strengths}
\end{table}

These objectives constrain only selected APD statistics. All unconstrained
features continue to be determined by the pretrained flow. Consequently, the
method provides approximate training-free control rather than exact sampling
from a conditional distribution, and sufficiently strong or conflicting
objectives may move samples away from the learned data manifold.

\section{Anisotropy matrix parametrisation}
\label{app:anisotropy}

The anisotropy matrix \(A_i\) defining each grain must be symmetric positive definite (SPD). Directly predicting the entries of \(A_i\) with an unconstrained neural network would not preserve this constraint and could therefore produce parameters that do not define a valid anisotropic norm. We instead construct an unconstrained, reversible parametrisation in which any network output corresponds to a valid anisotropy matrix.

Since \(A_i\) is SPD, its matrix logarithm
\[
    S_i = \log A_i
    = Q_i \log(\Lambda_i) Q_i^\top
\]
is real and symmetric, where
\[
    A_i = Q_i \Lambda_i Q_i^\top
\]
is the eigendecomposition of \(A_i\). Conversely, exponentiating any real symmetric matrix produces an SPD matrix.

We additionally separate anisotropic shape from isotropic scale by constraining
\[
    \det A_i = 1.
\]
The overall size of the corresponding grain is controlled separately through the APD weights, while this constraint allows \(A_i\) to represent only anisotropic shape.
Using the identity
\[
    \operatorname{tr}(\log A_i)=\log\det A_i,
\]
the unit-determinant constraint implies
\[
    \operatorname{tr}(S_i)=0.
\]
Hence \(S_i\) is a symmetric trace-free \(2\times2\) matrix and can be written using only two unconstrained scalars,
\[
    S_i =
    \begin{pmatrix}
        a_{i,1} & a_{i,2}\\
        a_{i,2} & -a_{i,1}
    \end{pmatrix},
    \qquad
    a_i=(a_{i,1},a_{i,2})\in\mathbb{R}^2.
\]
The inverse transformation is simply
\[
    A_i = \exp(S_i).
\]
Since the exponential of a symmetric matrix is SPD and
\[
    \det(\exp S_i)
    =
    \exp(\operatorname{tr}S_i)
    =
    1,
\]
every \(a_i\in\mathbb{R}^2\) corresponds to a valid unit-determinant anisotropy matrix. The network can therefore predict \(a_i\) without constraints while validity of the resulting APD parametrisation is preserved by construction.

This representation also gives a simple transformation rule under rotations. If \(R\) is a rotation of the spatial domain, then
\[
    A_i' = R A_i R^\top
    \qquad\Longrightarrow\qquad
    S_i' = R S_i R^\top.
\]
For a \(90^\circ\) rotation, this gives
\[
    a_i'=-a_i,
\]
while a \(180^\circ\) rotation leaves \(a_i\) unchanged. This transformation law is used in the construction of the \(C_4\)-equivariant velocity field below.

\section{\texorpdfstring{\(C_4\)-equivariant model updates}{C4-equivariant model updates}}
\label{app:equivariant-updates}

Before training, the APD parameters are transformed to numerically convenient coordinates. Seed positions originally defined on \([0,1]^2\) are mapped to
\[
    \widetilde{x}_i = 2x_i-1 \in [-1,1]^2.
\]
Besides placing the positional variables on a scale comparable to the Gaussian source distribution, this centres the square domain at the origin. Consequently, rotations about the centre of the physical domain act linearly in the transformed coordinates,
\[
    \widetilde{x}_i \mapsto R\widetilde{x}_i,
    \qquad R\in C_4.
\]
The scalar APD weights are standardised using statistics computed from the training set,
\[
    \widetilde{w}_i
    =
    \frac{w_i-\mu_w}{\sigma_w}.
\]
Since the weights are invariant under spatial rotations, this transformation does not affect their \(C_4\) transformation law.\\
The anisotropy parameters \(a_i\) are left centred at the origin so that their transformation law under \(C_4\) remains \(a_i\mapsto -a_i\) for a \(90^\circ\) rotation. The source state at $t=0$ is sampled from a Gaussian distribution. Generated states are transformed back to the original APD parametrisation before rasterisation.
For notational simplicity, we omit tildes for normalised quantities throughout this section.

Our objective is to construct a velocity field satisfying
\[
    v_\theta(R\mathbf{z},t)
    =
    R\,v_\theta(\mathbf{z},t)
\]
for every rotation \(R\in C_4\), where \(C_4\) contains rotations by multiples of \(90^\circ\). Here the action of \(R\) on \(\mathbf{z}\) includes the corresponding transformations of grain positions and anisotropy parameters, while the scalar weights \(w_i\) remain invariant.

A standard multilayer perceptron does not in general preserve equivariance when applied directly to orientation-dependent geometric quantities. Drawing inspiration from \citet{satorras_en_2022}, we therefore separate the update into two components. First, the message-passing network operates on \(C_4\)-invariant scalar features and predicts invariant scalar coefficients. Second, these coefficients multiply fixed geometric basis vectors with known transformation properties. The learned part of the model therefore determines the magnitude of each contribution, while the geometric bases determine how that contribution transforms under rotation.

For each directed edge \(e=(s,r)\), connecting sender \(s\) to receiver \(r\), define the displacement
\[
    d_e=x_r-x_s
\]
and the quadratic map
\[
    \rho(d_e)
    =
    \bigl(
        d_{e,1}^2-d_{e,2}^2,\,
        2d_{e,1}d_{e,2}
    \bigr).
\]
For a two-dimensional vector \(a=(a_1,a_2)\), we also define
\[
    a^\perp=(-a_2,a_1).
\]

The message-passing network receives invariant geometric information associated with each edge given by
\[
\phi_{e}=
\big[
 \|d_{e}\|^2,\,
 w_s,w_r,\,
 \|a_s\|^2,\|a_r\|^2,\,
 a_s^\top a_r,\,
 a_s\times a_r,\,
 a_s^\top\rho(d_{sr}),\,
 a_r^\top\rho(d_{sr}),\,
 b(x_s),b(x_r)
\big]\in\mathbb R^{15},
\]
where for \(u=(u_1,u_2)\) and \(v=(v_1,v_2)\), we define
\(
u\times v=u_1v_2-u_2v_1\), and \(
\rho(u)=(u_1^2-u_2^2,\,2u_1u_2);
\)
while
\(
b(u)=(u_1^2u_2^2,\,u_1^2+u_2^2,\,u_1^4+u_2^4)
\)
encodes position relative to the square boundary while remaining
\(C_4\)-invariant.

From the resulting edge representation, the output head predicts eleven invariant scalar coefficients,
\[
    \alpha_{e,1},\ldots,\alpha_{e,4},
    \qquad
    \beta_{e,1},\ldots,\beta_{e,6},
    \qquad
    \omega_e.
\]
An additional invariant attention score is normalised over the incoming edges of each receiver to obtain \(\gamma_e\). The attention weight allows the model to control the relative contribution of neighbouring grains while preserving equivariance.

The contribution of edge \(e\) to the positional velocity is
\[
    m_e^x
    =
    \gamma_e
    \left(
        \alpha_{e,1} d_e
        +\alpha_{e,2} x_s
        +\alpha_{e,3} S_s d_e
        +\alpha_{e,4} S_r d_e
    \right).
\]
Here \(S_i\) are the symmetric trace-free \(2\times2\) matrices, as in Appendix~\ref{app:anisotropy}.  The vectors
\[
    d_e,\qquad x_s,\qquad S_s d_e,\qquad S_r d_e
\]
transform in the same way as spatial positions under \(C_4\). Since their coefficients and \(\gamma_e\) are invariant, \(m_e^x\) transforms equivariantly under rotations.

The contribution to the anisotropy velocity is
\[
    m_e^a
    =
    \gamma_e
    \left(
        \beta_{e,1}a_s
        +\beta_{e,2}a_r
        +\beta_{e,3}\rho(d_e)
        +\beta_{e,4}a_s^\perp
        +\beta_{e,5}a_r^\perp
        +\beta_{e,6}\rho(d_e)^\perp
    \right).
\]
These bases transform according to the same representation as the anisotropy parameters. In particular, under a \(90^\circ\) spatial rotation,
\[
    a_i\mapsto -a_i
\]
and
\[
    \rho(Rd_e)=-\rho(d_e).
\]
The perpendicular quantities transform in the same way. Hence \(m_e^a\) has the required transformation law for an anisotropy velocity.

Finally, since the APD weight \(w_i\) is rotationally invariant, its edge contribution is simply
\[
    m_e^w
    =
    \gamma_e\omega_e.
\]

The node velocities are obtained by aggregating all incoming edge contributions,
\[
    v_{x_r}
    =
    \sum_{e:\,r(e)=r} m_e^x,
    \qquad
    v_{a_r}
    =
    \sum_{e:\,r(e)=r} m_e^a,
    \qquad
    v_{w_r}
    =
    \sum_{e:\,r(e)=r} m_e^w.
\]
Summation preserves the transformation properties of the individual messages. Since all learned coefficients are functions only of \(C_4\)-invariant features, while each geometric basis transforms according to the required representation, the complete vector field satisfies
\[
    v_\theta(R\mathbf{z},t)
    =
    R\,v_\theta(\mathbf{z},t),
    \qquad R\in C_4.
\]
Consequently, numerical integration of the flow preserves this equivariance: rotating the initial state by any multiple of \(90^\circ\) produces the corresponding rotation of the generated APD.

% \newpage
% \input{checklist.tex}

\end{document}